\documentclass[10pt,twocolumn,letterpaper]{article}

\usepackage[top=1in,bottom=1in,left=0.75in,right=0.75in]{geometry}
\usepackage{times}
\usepackage{booktabs}
\usepackage{multirow}
\usepackage{amsmath}
\usepackage{amssymb}
\usepackage{amsthm}
\usepackage{graphicx}
\usepackage{algorithm}
\usepackage{algorithmic}
\usepackage{xcolor}
\usepackage[numbers]{natbib}
\usepackage{hyperref}
\usepackage{subcaption}
\usepackage{tikz}
\usetikzlibrary{shapes.geometric, arrows.meta, positioning, calc, decorations.pathreplacing, fit, backgrounds}
\usepackage{caption}
\newtheorem{definition}{Definition}

\newcommand{\method}{\textsc{SalesLoop}}

\newcommand{\RR}{\mathbb{R}}
\newcommand{\EE}{\mathbb{E}}

\newcommand{\LL}{\mathcal{L}}
\newcommand{\DD}{\mathcal{D}}
\newcommand{\XX}{\mathcal{X}}
\newcommand{\YY}{\mathcal{Y}}
\newcommand{\BB}{\mathcal{B}}

\title{SalesLoop: Reinforcement Learning from Performance Feedback\\for Sales Lead Ranking}
\author{
  Chenyu Zhang \\
  Li Auto Inc. \\
  \texttt{cyzhang57@outlook.com}
}
\date{}

\hypersetup{
    colorlinks=true,
    linkcolor=blue,
    citecolor=blue,
    urlcolor=blue,
    pdftitle={SalesLoop: Reinforcement Learning from Performance Feedback for Sales Lead Ranking},
    pdfauthor={Chenyu Zhang}
}

\begin{document}
\maketitle

\begin{abstract}
Lead ranking in Customer Relationship Management (CRM) systems faces a persistent challenge: models achieving high offline accuracy often underperform in production. We identify three fundamental gaps responsible for this disconnect: offline-online metric mismatch, pointwise-listwise objective misalignment, and temporal distribution drift. To address these gaps, we propose SalesLoop, a reinforcement learning framework that establishes a closed feedback loop between model predictions and real-world business outcomes. Our approach introduces (1) a performance-aware reward that encodes conversion outcomes weighted by ranking position and conversion velocity, and (2) Discriminative GRPO, a listwise optimization objective that adapts Group Relative Policy Optimization to discriminative ranking models.

SalesLoop improves NDCG@K by +7.9\% and P@K by +15.8\% over the strongest static baseline. A 160-day production A/B test at a New Energy Vehicle manufacturer, spanning 16.5M leads and 280 sales specialists across two provincial markets, validates statistically significant cumulative lift of +4.7\% ($p=0.047$) and +8.7\% ($p=0.002$). In production, the ranking backbone achieves Top-10\% recall of 44.1\% and surfaces high-intent leads at $2.3\times$ the conversion rate of specialist baselines.
\end{abstract}

\section{Introduction}
\label{sec:intro}

Lead ranking---prioritizing sales leads by conversion likelihood---is a cornerstone of modern Customer Relationship Management (CRM)~\cite{winer2001crm,payne2005strategic}.
In high-value industries such as automotive sales, real estate, and enterprise B2B, the quality of lead prioritization directly determines sales efficiency and revenue.
Industrial CRM platforms routinely process millions of leads daily, but limited sales capacity means that follow-up effort must be prioritized: only the top-ranked leads (e.g., the daily Top-10,000) are surfaced to specialists for timely follow-up.
Because the highest-priority effort is concentrated on these top-ranked leads, how well the model ranks them directly shapes how many conversions the sales team realizes---and thus revenue.

Despite significant advances in deep learning~\cite{guo2017deepfm,wang2017dcn} and the recent emergence of Large Language Models (LLMs) for CRM tasks~\cite{huang2024crmarena}, a persistent gap remains between model development and real-world deployment.
In our experience deploying lead ranking systems at a major New Energy Vehicle manufacturer, we observe that models achieving high offline metrics frequently underperform in production.
We identify three fundamental gaps that existing approaches fail to address:

\paragraph{Gap 1: Offline-Online Metric Mismatch.}
Offline model evaluation optimizes for prediction accuracy (e.g., AUC on binary conversion labels), but the online business outcome depends on a fundamentally different set of factors: the \textit{effectiveness of sales follow-up}.
In practice, conversion is not a passive event that a model merely predicts---it is an outcome jointly determined by lead quality \textit{and} the intensity of human effort (follow-up frequency, call duration, engagement depth).
This creates a critical evaluation mismatch: an offline metric can improve while the corresponding online outcome stagnates or even deteriorates.
An analogous phenomenon has been documented in multi-objective recommendation systems, where offline AUC improvements for both watch time and user interaction coexist, yet online deployment yields increased watch time \textit{alongside} decreased interaction~\cite{xu2025emer}.
The subtlety is that historical conversion labels are themselves confounded by human effort: a lead converts partly because it was promising and partly because a specialist invested time in it.
A model trained on these labels therefore learns a mixture of \textit{intent} and \textit{effort}, and can raise offline AUC by favoring leads that are ``easier to follow up'' (e.g., responsive but low-intent) over leads that need more effort but carry higher latent intent.
The result: offline AUC increases, but online lock-in rate does not, because the model's ranking no longer aligns with the effective deployment of limited sales capacity.

\paragraph{Gap 2: Pointwise Prediction $\neq$ Listwise Ranking Quality.}
Standard training objectives optimize pointwise accuracy: minimizing prediction error for individual leads.
However, business value depends entirely on \textit{listwise} ranking quality: whether the \textit{right} leads appear within the operational Top-$K$ window.
These two objectives are not equivalent.
A model may achieve excellent pointwise accuracy across the full lead population, yet still rank true conversions at positions 11K--20K---just outside the Top-10K operational window where sales capacity is allocated.
In this scenario, the model's pointwise performance is strong, but its business impact is zero, because the leads that actually convert are not surfaced to sales teams.
Conversely, a model with modest pointwise accuracy but strong Top-$K$ concentration of conversions delivers substantially higher business value.
Pointwise-optimal models are not necessarily listwise-optimal, and optimizing for the former does not guarantee improvement in the latter.

\paragraph{Gap 3: Temporal Distribution Drift.}
Customer behaviors, market conditions, and sales strategies continuously evolve over time~\cite{gama2014survey}.
Promotional campaigns, seasonal variations, and competitive dynamics cause the lead population distribution to shift, yet static models trained on historical data remain fixed.
In automotive sales, the Chinese New Year promotional period alone can shift lead volume by 40--60\%, with qualitatively different conversion patterns (e.g., higher volume of casual browsers, lower immediate intent).
A model that performed well in one quarter may become progressively misaligned a few months later because the underlying distribution of incoming leads has changed.
Traditional training pipelines lack mechanisms for continuous adaptation to these temporal shifts, creating a growing gap between what the model was trained on and what it encounters in production.

\medskip

A common thread runs through these three gaps: models are trained once on historical data but must operate in a continuously evolving deployment environment. Addressing them calls for a system that adapts to feedback from production rather than relying solely on historical signals.
Inspired by recent advances in Reinforcement Learning from Performance Feedback (RLPF) for ad optimization~\cite{jiang2025adtext} and Group Relative Policy Optimization (GRPO) for language models~\cite{shao2024deepseekmath}, we propose \textbf{\method{}}, a reinforcement learning framework that closes the feedback loop between model deployment and business outcomes.

The reward signal comes directly from what deployment reveals: which leads actually converted, how quickly, and at what rank the model had placed them.
\method{} turns this deployment feedback into a continuous learning signal:
\begin{center}
\textit{Deploy $\rightarrow$ Observe $\rightarrow$ Reward $\rightarrow$ Update $\rightarrow$ Deploy}
\end{center}

Our approach introduces two key innovations:

(1) A \textbf{performance-aware reward} that encodes conversion outcomes weighted by ranking position (following the logarithmic attention decay model~\cite{chuklin2015click}) and conversion velocity (faster conversions indicate stronger intent).
Unlike binary conversion labels, this reward differentiates among converted leads and directly encodes ranking quality from deployment.

(2) A \textbf{listwise optimization objective} that adapts GRPO to discriminative ranking models---a setting we term \textit{Discriminative GRPO}.
By treating each training batch as a group and computing relative advantages across leads, we optimize the ranking distribution rather than individual predictions, directly addressing Gap 2.

We evaluate \method{} through a multi-stage protocol at a New Energy Vehicle manufacturer, anchored by a live 160-day production deployment spanning 16.5M leads and 280 sales specialists:

\begin{itemize}
    \item \textbf{Offline benchmarking}: \method{} achieves significant improvements over traditional ML (XGBoost, DeepFM) and LLM-based baselines (SFT, DPO), with the largest gains on ranking-sensitive metrics (NDCG@K: +7.9\%, P@K: +15.8\% over the strongest static baseline).
    \item \textbf{Production A/B test}: A 160-day deployment across two provincial markets (280 specialists) validates +4.7\% and +8.7\% cumulative lift in lock-in conversions ($p<0.05$), with the advantage growing as the feedback loop accumulates deployment data.
    \item \textbf{Deployment validation}: In production over 103 days, the ranking backbone attains Top-10\% recall of 44.1\% (4.4$\times$ random) and surfaces incremental high-intent leads at 2.3$\times$ the conversion rate of specialist baselines, quantifying absolute ranking quality and business value independent of the A/B comparison.
\end{itemize}

Our contributions are threefold:
\begin{enumerate}
    \item We characterize three gaps that cause offline-strong lead ranking models to underperform in production---offline-online metric mismatch, pointwise-listwise objective misalignment, and temporal distribution drift---and propose closing them with a performance-feedback loop rather than a static training pipeline.
    \item We instantiate this loop with two components: a performance-aware reward that combines conversion outcomes with ranking position and conversion velocity, and \textit{Discriminative GRPO}, a listwise objective that adapts group-relative advantages to discriminative ranking models.
    \item We validate \method{} in a 160-day production A/B test over 16.5M leads and 280 specialists across two markets---a long-horizon, real-world evaluation that is rare in online ranking---demonstrating statistically significant lift (+4.7\% to +8.7\%, $p<0.05$).
\end{enumerate}

\section{Related Work}
\label{sec:related}

\paragraph{Lead Scoring and CRM Analytics.}
Traditional lead scoring relies on rule-based scorecards~\cite{winer2001crm} or shallow machine learning models such as logistic regression and gradient boosting~\cite{chen2016xgboost}.
Recent work explores deep learning for CRM and click-through prediction, including Wide \& Deep~\cite{cheng2016wide}, DeepFM~\cite{guo2017deepfm}, xDeepFM~\cite{lian2018xdeepfm}, AutoInt~\cite{song2019autoint}, and DCN and its successor~\cite{wang2017dcn,wang2021dcn} for modeling feature interactions.
CRMArena~\cite{huang2024crmarena} benchmarks LLMs on CRM tasks including lead qualification.
SalesRLAgent~\cite{salesrlagent2025} applies RL to real-time conversion prediction in sales conversations.
Sun~et~al.~\cite{sun2025asllr} propose asLLR, integrating CTR and QA losses within a decoder-only LLM for automotive lead ranking, with gains validated in A/B testing.
HPRO~\cite{zhang2026rethinking} scores leads with an LLM under a hierarchical preference-ranking objective, but is trained once on historical data and deployed as a static model.
\method{} instead closes the loop between deployment and training, continuously adapting the ranking model from observed conversion outcomes.

\paragraph{Learning to Rank.}
Learning-to-rank methods are categorized into pointwise~\cite{liu2009ltr}, pairwise~\cite{burges2005ranknet}, and listwise~\cite{cao2007listnet,xia2008listwise} approaches.
Pointwise methods treat ranking as regression or classification on individual items.
Pairwise methods optimize relative orderings between item pairs.
Listwise methods directly optimize ranking quality over the entire list.
While listwise approaches better align with ranking objectives, most operate in offline settings with static relevance labels.
Online learning-to-rank~\cite{oosterhuis2018duoltr,jagerman2019osltr} adapts rankings from user feedback, but assumes immediate click signals rather than the delayed, sparse conversion outcomes with long feedback cycles (30+ days) that characterize lead ranking.

\paragraph{Reinforcement Learning for Ranking and Recommendation.}
RL has been applied to ranking and recommendation tasks~\cite{ie2019slaterl,chen2019topk}.
RLHF~\cite{ouyang2022rlhf} aligns LLMs using human feedback as a reward signal, with PPO~\cite{schulman2017ppo} as the standard policy-optimization algorithm.
Xue~et~al.~\cite{xue2025auro} apply RL to adaptive user retention in recommender systems, showing that online RL can outperform static scoring in production; their reward, however, is driven by near-immediate user-retention signals, whereas lead ranking must learn from conversions that materialize weeks later.
Most relevant to our work, Jiang~et~al.~\cite{jiang2025adtext} introduce Reinforcement Learning from Performance Feedback (RLPF) for ad text generation at Meta, using a CTR prediction model as the reward signal and reporting consistent CTR gains in a large-scale production A/B test.
This line of work establishes that real-world performance metrics can serve as effective reward signals, bypassing the need for human annotation.
Their setting, however, is \textit{generative}---optimizing token-level generation probabilities against an immediate CTR proxy.
GRPO~\cite{shao2024deepseekmath} likewise operates on generative models, introducing group-relative advantages that remove the need for a learned value network.
\method{} adapts these insights to a \textit{discriminative} ranking model under \textit{delayed, sparse} rewards, proposing a listwise variant that computes group-relative advantages over lead scores rather than token-level generation probabilities.

\paragraph{Online and Continual Learning.}
Online learning~\cite{shalev2012online} updates models incrementally as data arrives.
Continual learning~\cite{parisi2019continual} addresses catastrophic forgetting when learning from non-stationary distributions.
Concept drift detection~\cite{gama2014survey} identifies distribution shifts requiring model updates.
Recent work on unbiased learning to rank~\cite{schnabel2016recommendations,zhang2023disentangling} highlights the importance of disentangling relevance from position and selection bias in deployed ranking systems.
While these methods provide theoretical foundations, they rarely address the specific challenges of lead ranking: extremely delayed feedback (30+ days), sparse positive labels ($<$2\% conversion rate), and the need to optimize listwise business metrics rather than pointwise accuracy.

Taken together, prior work leaves a distinct gap: static offline LTR does not adapt after deployment; online LTR assumes immediate feedback; and performance-feedback RL has so far targeted generative models with near-term proxies. None addresses discriminative, listwise ranking that must learn from delayed and sparse business outcomes---the regime \method{} is designed for.

\section{Methodology}
\begin{figure*}[t]
\centering
\includegraphics[width=1.0\textwidth]{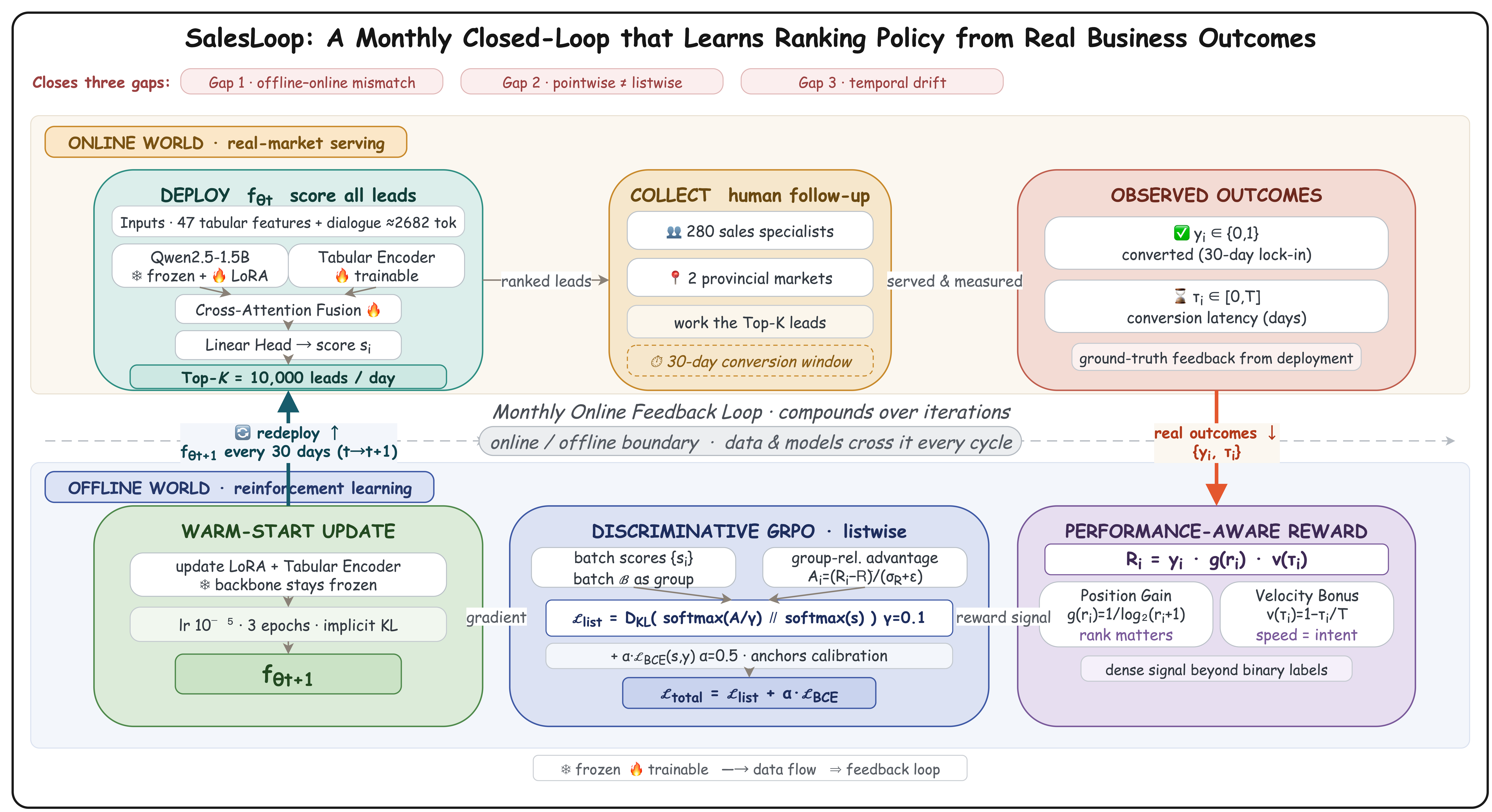}
\caption{Overview of the \method{} framework. In each monthly iteration, the deployed ranking model scores all leads and selects Top-$K$ candidates for sales follow-up. After the 30-day conversion window, observed outcomes are converted into performance-aware rewards, which are used to update the model through a Discriminative GRPO objective. The updated model is then deployed in the next iteration, forming a closed online feedback loop.}
\label{fig:framework}
\end{figure*}
\subsection{Problem Formulation}

Let $\XX$ denote the feature space and $\YY = \{0, 1\}$ the binary conversion outcomes.
At iteration $t$, we observe a deployment dataset $\DD_t = \{(\mathbf{x}_i, y_i, r_i, \tau_i)\}_{i=1}^{N_t}$, where:
\begin{itemize}
    \item $\mathbf{x}_i \in \XX$: lead features (tabular attributes + dialogue transcripts),
    \item $y_i \in \YY$: binary conversion indicator (1 if locked-in within $T$ days),
    \item $r_i \in \{1, \ldots, N_t\}$: rank assigned by the deployed model $f_{\theta_{t-1}}$,
    \item $\tau_i \in [0, T]$: conversion latency in days from ranking to lock-in, defined only for $y_i = 1$.
\end{itemize}

\noindent\textbf{Objective.}
Learn a scoring function $f_\theta: \XX \to \RR$ that maximizes business-relevant ranking metrics within the operational capacity $K$:
\begin{equation}
\label{eq:objective}
\max_\theta \; \EE_{(\mathbf{x}, y) \sim p_t}\left[ \text{Precision@}K + \lambda \cdot \text{Recall@}K \right],
\end{equation}
where $K$ is the operational capacity (e.g., $K=10{,}000$ leads for daily follow-up) and $\lambda$ expresses the precision-recall trade-off inherent in the business goal.
Eq.~\ref{eq:objective} states this goal conceptually; since Precision@$K$ and Recall@$K$ are non-differentiable, we do not optimize it directly (and thus do not tune $\lambda$ explicitly). Instead, our position-weighted listwise objective (\S\ref{sec:optimization}) serves as a differentiable surrogate that concentrates converted leads near the top of the ranking, which is precisely what Top-$K$ metrics reward.
The key challenge is that $p_t$---the deployment distribution at time $t$---differs from historical training distributions and evolves over time due to market dynamics and seasonal patterns.

\subsection{SalesLoop Framework}

\method{} operates as a closed-loop system with three stages per iteration (Figure~\ref{fig:framework}):

\noindent\textbf{Stage 1: Deploy.}
Model $f_{\theta_t}$ scores all incoming leads and produces a ranking.
The Top-$K$ leads are selected for sales follow-up based on operational capacity constraints.

\noindent\textbf{Stage 2: Collect.}
After the conversion window $T$ (30 days), we observe real-world outcomes: which leads converted ($y_i$) and how quickly ($\tau_i$).
These outcomes are transformed into performance-aware rewards $R_i$ (\S\ref{sec:reward}).

\noindent\textbf{Stage 3: Update.}
The model is optimized via a listwise loss function that aligns predicted scores with observed rewards (\S\ref{sec:optimization}).
Training continues from $f_{\theta_t}$ (warm start) with a reduced learning rate to preserve previously learned representations while adapting to new patterns.

This closed-loop cycle repeats monthly, enabling the model to continuously adapt to the evolving deployment distribution.
A critical design choice is the warm-start strategy: rather than training from scratch, each iteration initializes from the previous model's parameters, providing a strong prior that accelerates convergence while the reduced learning rate and regularization keep the model close to previously learned patterns.
This design enables \method{} to accumulate knowledge across iterations, creating a compounding effect where each iteration benefits from all prior deployment experience.

\subsection{Performance-Aware Reward Design}
\label{sec:reward}

The reward design is central to \method{}'s ability to optimize ranking quality rather than pointwise accuracy.
We construct rewards that encode three business-relevant signals: the conversion outcome ($y_i$), the ranking position at which the lead was surfaced ($r_i$), and the speed of conversion ($\tau_i$).

\begin{definition}[Performance-Aware Reward]
\label{def:reward}
For lead $i$ with conversion outcome $y_i \in \{0,1\}$, assigned rank $r_i$, and conversion latency $\tau_i$:
\begin{equation}
\label{eq:reward}
R_i = y_i \cdot g(r_i) \cdot v(\tau_i),
\end{equation}
where the position gain $g(r_i)$ and velocity bonus $v(\tau_i)$ are defined as:
\begin{equation}
g(r_i) = \frac{1}{\log_2(r_i + 1)}, \qquad v(\tau_i) = 1 - \frac{\tau_i}{T}.
\end{equation}
The leading factor $y_i$ ensures $R_i = 0$ for non-converted leads, so $v(\tau_i)$ only needs to be defined for converted leads, where $\tau_i \in [0, T]$.
\end{definition}

\noindent\textbf{Position gain.}
We adopt logarithmic decay following the position-based click model~\cite{chuklin2015click}, which assumes user (sales) attention decays logarithmically with rank.
This choice is motivated by our operational setting: a conversion at rank 1 is substantially more valuable than at rank 10{,}000, as the former represents a correctly prioritized lead while the latter indicates a missed opportunity.
A conversion at rank 1 yields $g(1)=1.0$; at rank 1{,}000, $g(1000) \approx 0.1$.

\noindent\textbf{Velocity bonus.}
Faster conversions indicate stronger purchase intent.
We use linear decay over the conversion window $T$: leads that convert on day 1 receive the maximum velocity bonus ($v=1$), while those converting on day 30 receive zero ($v=0$).
This simple, interpretable signal differentiates among converted leads---something binary labels cannot do.

\noindent\textbf{Reward properties.}
The reward $R_i$ is zero for non-converted leads and strictly positive for converted leads, with magnitude determined by ranking position and conversion speed.
This design ensures that: (1) the model receives no credit for non-conversions, (2) correctly ranking high-value leads (early in the list) yields higher rewards, and (3) faster conversions are preferentially rewarded, encouraging the model to identify leads with immediate purchase intent.

\subsection{Discriminative GRPO for Ranking}
\label{sec:optimization}

We adapt Group Relative Policy Optimization (GRPO)~\cite{shao2024deepseekmath}---originally designed for generative language models---to \textbf{discriminative ranking models}.
This adaptation, which we term \textit{Discriminative GRPO}, preserves the core benefit of GRPO (variance reduction through group-normalized advantages) while enabling its application to scoring functions.

\paragraph{From generative to discriminative.}
In the original GRPO framework, the model generates multiple candidate responses for each prompt, and advantages are computed across candidates within each prompt group.
In our discriminative setting, the model outputs a scalar score $s_i = f_\theta(\mathbf{x}_i)$ for each lead.
We treat each training batch as a ``group'' and compute relative advantages across leads within the batch.
This parallels the generative case: instead of comparing different text outputs for the same input, we compare different lead scores within the same batch.

\paragraph{Scope of the adaptation.}
We emphasize which parts of GRPO we adopt and which we do not.
We inherit GRPO's central idea---replacing a learned value baseline with \textit{group-relative}, within-batch advantage normalization---which is what stabilizes learning under our sparse rewards.
We deliberately do \textit{not} carry over the components tied to autoregressive generation: there is no importance-sampling ratio $\pi_\theta/\pi_{\theta_\text{old}}$ and no PPO-style probability-ratio clipping.
These are unnecessary here because a discriminative ranker produces a single scalar per lead rather than a sampled token sequence, so there is no per-token, off-policy credit-assignment problem to correct.
Instead, we regulate policy drift across iterations through the explicit BCE regularization of Eq.~\ref{eq:total} and the conservative warm-start learning rate described in the implementation details.
We therefore view Discriminative GRPO as a group-relative, reward-weighted listwise objective in the spirit of GRPO, rather than a literal transplant of the PPO-based algorithm.

\paragraph{Listwise optimization.}
For a training batch $\BB = \{(\mathbf{x}_i, R_i)\}_{i=1}^{B}$:

\textbf{Step 1: Compute model scores.}
\begin{equation}
s_i = f_\theta(\mathbf{x}_i), \quad i = 1, \ldots, B.
\end{equation}

\textbf{Step 2: Compute group-relative advantages.}
Standardize rewards within batch to obtain relative advantages:
\begin{equation}
\label{eq:advantage}
A_i = \frac{R_i - \bar{R}}{\sigma_R + \epsilon}, \quad \bar{R} = \frac{1}{B}\sum_{j=1}^{B} R_j, \quad \sigma_R = \sqrt{\frac{1}{B}\sum_{j=1}^{B}(R_j - \bar{R})^2},
\end{equation}
where $\epsilon = 10^{-8}$ for numerical stability.
This standardization removes the batch-level mean and scales by variance, analogous to GRPO's group normalization.
In sparse-reward settings ($\sim$1.5\% conversion rate in our deployment), this reduces variance by centering around the batch average rather than an absolute baseline, which is critical for stable gradient estimation when the majority of labels are zero.

\textbf{Step 3: Align distributions via KL divergence.}
We optimize the score distribution to match the advantage distribution:
\begin{equation}
\label{eq:listloss}
\LL_{\text{list}} = D_{\text{KL}}\Big( \text{softmax}(\mathbf{A} / \gamma) \;\Big\|\; \text{softmax}(\mathbf{s}) \Big),
\end{equation}
where $\gamma > 0$ is a temperature controlling the sharpness of the target distribution.
The two operations are complementary rather than redundant: the standardization in Eq.~\ref{eq:advantage} fixes the \textit{scale} of advantages across batches, while $\gamma$ controls how sharply probability mass concentrates on the highest-reward leads.
Lower $\gamma$ creates sharper targets that emphasize high-reward leads; higher $\gamma$ produces softer targets.
We set $\gamma = 0.1$ based on sensitivity analysis (\S\ref{sec:sensitivity}).

This formulation is listwise because the softmax couples all leads in the batch: changing one lead's score affects its relative probability mass, directly optimizing the ranking distribution rather than individual predictions.

\paragraph{Relation to ListNet.}
The listwise loss in Eq.~\ref{eq:listloss} is structurally related to ListNet~\cite{cao2007listnet}, which also matches score and target distributions via a cross-entropy/KL objective.
The key difference is the source of the target: ListNet uses static relevance labels, whereas our target distribution is built from \textit{dynamic} performance-aware rewards that encode real deployment feedback (position and velocity), combined with the group-relative normalization inherited from GRPO.

\paragraph{Training objective.}
We combine the listwise loss with pointwise regularization to prevent calibration drift:
\begin{equation}
\label{eq:total}
\LL_{\text{total}} = \LL_{\text{list}} + \alpha \cdot \LL_{\text{BCE}}(\mathbf{s}, \mathbf{y}),
\end{equation}
where $\LL_{\text{BCE}}$ is binary cross-entropy against conversion labels and $\alpha$ controls regularization strength.
The pointwise term anchors individual predictions to conversion labels, while the listwise term optimizes the relative ordering.
We set $\alpha = 0.5$ based on ablation studies.

\subsection{Implementation Details}

\paragraph{Model architecture.}
We use Qwen2.5-1.5B~\cite{qwen2024} as the language backbone for processing dialogue transcripts, augmented with a tabular encoder for structured features.
Both representations are fused via cross-attention and projected to a scalar score through a linear head.
This architecture follows our prior work on HPRO~\cite{zhang2026rethinking}.

\paragraph{Efficient fine-tuning.}
We apply LoRA~\cite{hu2022lora} with rank $r=16$ and scaling factor $\alpha=32$, targeting query and value projection layers.
The tabular encoder and fusion layers are fully trained.
Online iterations use a reduced learning rate of $10^{-5}$ (10$\times$ lower than initial SFT) with batch size 64 for 3 epochs.
This conservative update strategy serves two purposes: (1) it preserves the base model's calibration while adapting to distributional shifts, and (2) it provides an implicit KL constraint between consecutive iterations, complementing the explicit regularization in Eq.~\ref{eq:total}.

\paragraph{Hyperparameters.}
Temperature $\gamma = 0.1$, regularization $\alpha = 0.5$, conversion window $T = 30$ days.
Iteration frequency: monthly, aligned with the 30-day conversion window.
Initial model $f_{\theta_0}$ is trained via supervised fine-tuning (SFT) on 9.2M historical samples (July--September 2025) with binary cross-entropy loss.
Optimization uses AdamW~\cite{loshchilov2019adamw} with linear warmup (5\% of steps) and cosine decay.

\section{Experiments}

We evaluate \method{} through a comprehensive experimental protocol spanning offline benchmarking, long-term online deployment across two provincial markets, and detailed mechanism analysis.

\subsection{Experimental Setup}

\paragraph{Datasets.}
We collect lead data from a leading New Energy Vehicle (NEV) manufacturer's CRM system.
Our evaluation uses two datasets spanning different operational periods:

\begin{itemize}
    \item \textbf{Historical Dataset} (9.2M samples, $\sim$1.5\% positive): Collected from July--September 2025. Used for initial model training and fair comparison with offline baselines.
    \item \textbf{Online Deployment Dataset} (16.5M samples, $\sim$1.5\% positive): Collected from December 2025--June 2026, representing a 160-day production A/B test with delayed conversion feedback across two provincial markets.
\end{itemize}

In addition, we use a 103-day production log (September 20 -- December 31, 2025) from the same CRM system to characterize the ranking quality of the deployed base model prior to the A/B test (\S\ref{sec:ranking_quality}).

Each lead comprises 47 tabular features (demographics, source channel, interaction history, behavioral signals) and sales dialogue transcripts (average 2,682 tokens).
The operational capacity $K=10{,}000$ represents the top-ranked leads surfaced to sales specialists daily for prioritized follow-up.
Conversion labels are determined by 30-day lock-in status.
Table~\ref{tab:dataset} summarizes statistics.

\begin{table}[t]
\centering
\caption{Dataset statistics for lead ranking evaluation.}
\label{tab:dataset}
\small
\begin{tabular}{lc}
\toprule
\textbf{Statistic} & \textbf{Value} \\
\midrule
Total leads (Historical) & 9.2M \\
Total leads (Online, 160 days) & 16.5M \\
Tabular features & 47 \\
Avg. dialogue tokens & 2,682 \\
Conversion window $T$ & 30 days \\
Operational capacity $K$ & 10,000 \\
\bottomrule
\end{tabular}
\end{table}

\paragraph{Baselines.}
We compare against representative methods spanning traditional ML, deep learning, and LLM-based approaches:
\begin{itemize}
    \item \textbf{XGBoost}~\cite{chen2016xgboost}: Gradient boosting on tabular features only, representing the traditional industrial baseline.
    \item \textbf{DeepFM}~\cite{guo2017deepfm}: Deep factorization machine capturing feature interactions, representing deep CTR models.
    \item \textbf{LLM+SFT}: Qwen2.5-1.5B fine-tuned on historical conversion labels with binary cross-entropy loss. This serves as our base model $f_{\theta_0}$ and the \textbf{control group} in online A/B testing.
    \item \textbf{LLM+DPO}~\cite{rafailov2024dpo}: Direct Preference Optimization using converted vs.\ non-converted lead pairs, representing static preference-based methods.
\end{itemize}

\paragraph{Metrics.}
We report standard ranking metrics at $K=10{,}000$:
\begin{itemize}
    \item \textbf{Precision@K (P@K)}: Fraction of Top-$K$ leads that convert (lock-in rate within Top-$K$).
    \item \textbf{Recall@K (R@K)}: Fraction of all conversions captured in Top-$K$.
    \item \textbf{NDCG@K}: Normalized discounted cumulative gain.
    \item \textbf{AUC}: Area under ROC curve for pointwise prediction accuracy.
\end{itemize}

\paragraph{Implementation.}
All LLM-based methods share the same architecture: Qwen2.5-1.5B with LoRA ($r=16$, $\alpha=32$) for efficient fine-tuning, targeting query and value projections.
The tabular encoder and cross-attention fusion layers are fully trained.
Training uses AdamW with learning rate $10^{-5}$, batch size 64, for 3 epochs per iteration on 8$\times$A100 GPUs.
\method{} performs monthly online iterations on the deployment dataset, using the previous month's conversion outcomes as reward signals.

\paragraph{Online A/B Test Configuration.}
The production A/B test was conducted simultaneously in two anonymized provincial markets with the following configuration:

\begin{itemize}
    \item \textbf{Duration}: 160 days (December 26, 2025 -- June 4, 2026).
    \item \textbf{Province A}: 60 treatment specialists (\method{}) vs.\ 60 control specialists (LLM+SFT), randomly assigned.
    \item \textbf{Province B}: 80 treatment specialists (\method{}) vs.\ 80 control specialists (LLM+SFT), randomly assigned.
    \item \textbf{Primary metric}: Cumulative lock-in conversions and per-specialist productivity.
\end{itemize}

The control group uses the static LLM+SFT model ($f_{\theta_0}$) throughout the entire test period without any parameter updates.
The treatment group uses \method{}, which is initialized from the same $f_{\theta_0}$ but undergoes monthly online iterations using deployment feedback.

\subsection{Offline Benchmarking}
\label{sec:offline}

We first evaluate all methods under a controlled offline setting using the Historical Dataset, ensuring fair comparison with consistent data splits and temporal ordering.
Table~\ref{tab:main} presents results.

\begin{table*}[t]
\centering
\caption{Offline results on Historical Dataset ($K=10{,}000$). For a controlled, single-round comparison with the static baselines, \method{} is evaluated after 1 iteration using held-out feedback; the multi-iteration behavior is analyzed separately in the ablation (Table~\ref{tab:ablation}) and online study. Best in \textbf{bold}, second \underline{underlined}.}
\label{tab:main}
\small
\begin{tabular}{lccccc}
\toprule
\textbf{Method} & \textbf{AUC} & \textbf{P@K} & \textbf{R@K} & \textbf{NDCG@K} & \textbf{MRR} \\
\midrule
XGBoost & 0.7241 & 4.52\% & 5.03\% & 0.3017 & 0.0891 \\
DeepFM & 0.7389 & 5.04\% & 5.61\% & 0.3185 & 0.0934 \\
\midrule
LLM+SFT & \underline{0.7814} & 6.37\% & 7.08\% & 0.3674 & \underline{0.1102} \\
LLM+DPO & 0.7762 & \underline{6.53\%} & \underline{7.26\%} & \underline{0.3718} & 0.1064 \\
\midrule
\textbf{\method{}} & \textbf{0.7891} & \textbf{7.56\%} & \textbf{8.40\%} & \textbf{0.4012} & \textbf{0.1203} \\
\bottomrule
\end{tabular}
\end{table*}

\paragraph{LLM backbones substantially outperform traditional models.}
LLM+SFT achieves +40.9\% P@K over XGBoost (6.37\% vs.~4.52\%), demonstrating the value of dialogue understanding.
This aligns with our hypothesis: sales conversations contain rich intent signals that tabular features alone cannot capture.
However, the AUC gap between LLM+SFT and DeepFM (0.781 vs.~0.739) is notably smaller than the P@K gap (+26.4\%), suggesting that AUC alone underestimates the business value of LLM-based ranking---precisely because of the pointwise-listwise gap we identified (Gap 2).

\paragraph{DPO does not uniformly improve over SFT.}
LLM+DPO achieves slightly higher P@K and NDCG@K than LLM+SFT (+2.5\% and +1.2\% respectively), but its AUC is \textit{lower} (0.776 vs.~0.781) and MRR also drops.
This suggests that DPO's pairwise preference objective can hurt pointwise calibration while improving top-$K$ discrimination---a known trade-off between ranking and classification objectives~\cite{rafailov2024dpo}.

\paragraph{\method{} achieves the best performance across all metrics.}
Compared to the better of LLM+SFT and LLM+DPO on each metric, \method{} improves P@K by +15.8\% and NDCG@K by +7.9\%, while also recovering AUC to 0.789.
The consistent gains across both ranking-sensitive and pointwise metrics validate that our listwise optimization with performance-aware rewards effectively addresses Gap 2 without sacrificing calibration.

\subsection{Online Deployment Results}
\label{sec:online}

We now present the core empirical contribution of this work: a 160-day production A/B test across two provincial markets. To our knowledge, this is among the longest continuously-running online RL evaluations reported for industrial lead ranking.

\subsubsection{Overall Results}

Table~\ref{tab:ab_summary} summarizes the end-of-period results.

\begin{table}[t]
\centering
\caption{Online A/B test results after 160 days of deployment.}
\label{tab:ab_summary}
\small
\begin{tabular}{lcc}
\toprule
\textbf{Metric} & \textbf{Province A} & \textbf{Province B} \\
\midrule
Team size (per group) & 60 & 80 \\
Treatment conversions & 2{,}142 & 2{,}653 \\
Control conversions & 2{,}046 & 2{,}441 \\
Absolute lift & +96 & +212 \\
\textbf{Relative lift} & \textbf{+4.7\%} & \textbf{+8.7\%} \\
Per-specialist lift & +1.6 & +2.7 \\
\midrule
Weekly $t$-test ($p$-value) & 0.047 & 0.002 \\
\bottomrule
\end{tabular}
\end{table}

\method{} achieves statistically significant improvements in both markets: +4.7\% cumulative lift in Province A ($p=0.047$, weekly paired $t$-test) and +8.7\% in Province B ($p=0.002$).
The effect in Province B is strongly significant, while Province A, though significant at the 0.05 level, is more marginal---consistent with its longer warm-up period before the feedback loop takes effect (\S\ref{sec:temporal}).
In per-specialist terms, each sales specialist in the treatment group converts 1.6--2.7 additional leads over the 160-day period compared to their control counterparts.
Because each conversion corresponds to a high-value durable good, these per-specialist gains translate into substantial revenue impact at team scale.

\subsubsection{Temporal Dynamics: The Feedback Loop Effect}
\label{sec:temporal}

\begin{figure*}[t]
    \centering
    \includegraphics[width=\textwidth]{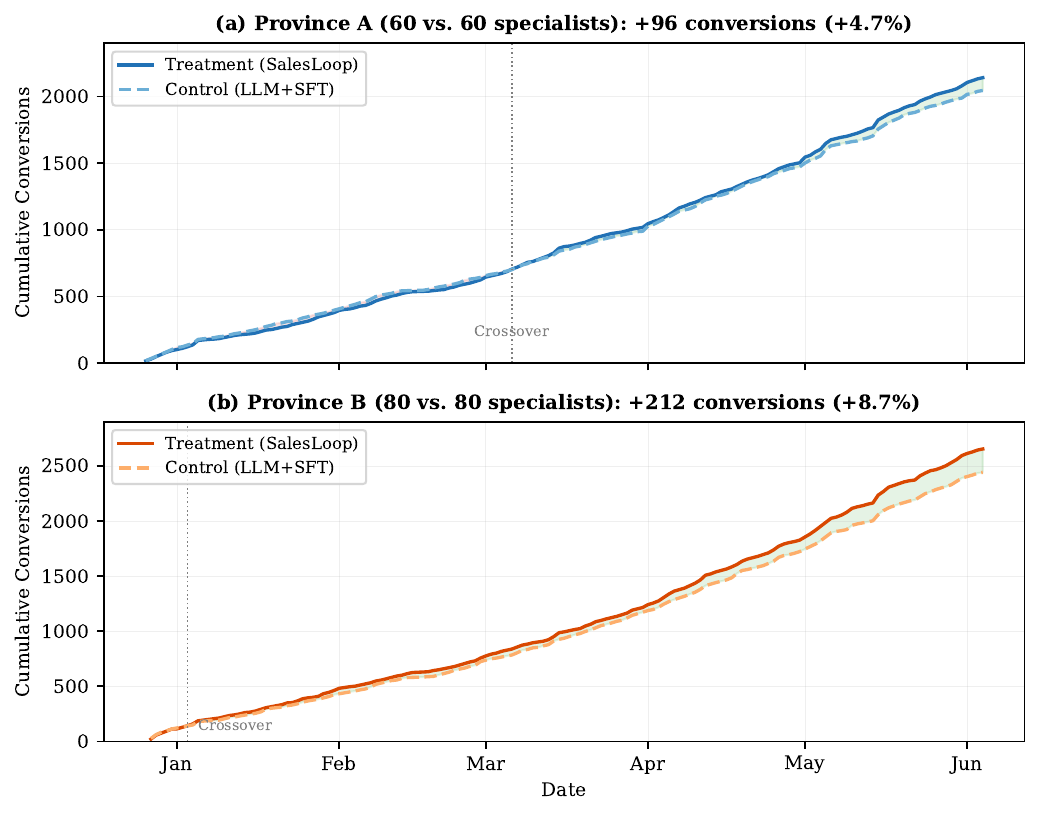}
    \caption{Cumulative lock-in conversions over 160 days. Green shading indicates periods where treatment outperforms control; red shading indicates the reverse. Vertical dashed lines mark the crossover point where \method{} begins to persistently outperform the static baseline.}
    \label{fig:cumulative}
\end{figure*}

Beyond the end-of-period totals, the \textit{temporal dynamics} of the treatment-control gap provide direct evidence for the feedback loop mechanism.
Figure~\ref{fig:cumulative} shows cumulative conversion curves for both provinces.

As deployment feedback accumulates, \method{} crosses over the static baseline and steadily widens the gap in both markets.
The cumulative difference grows from 0 to +96 (Province A) and +212 (Province B) by the end of the observation period, and once established the divergence does not reverse within our observation window---suggesting that the feedback loop yields a compounding advantage rather than a one-time boost.
The onset is fast: Province B turns positive within weeks, and the effect strengthens over successive iterations.
The main exception is an initial warm-up in Province A, where \method{} trails the control over roughly the first 70 days ($-5.9\%$ on average) before the loop takes effect---a one-time, first-iteration transient that can be avoided by warming up the initial iteration offline.
This difference in time-to-effect likely reflects market characteristics: Province B has a higher daily lead volume relative to team size, providing denser reward signals, which suggests that \method{}'s feedback loop efficiency is modulated by the \textit{information density} of the deployment environment.

\subsubsection{Month-over-Month Analysis}

\begin{figure}[t]
    \centering
    \includegraphics[width=\columnwidth]{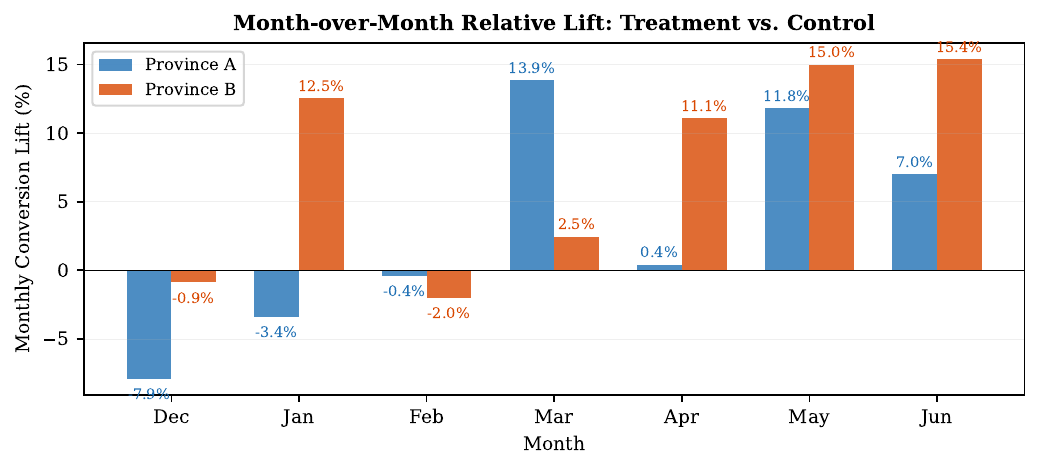}
    \caption{Monthly conversion lift (\%) of treatment over control. Province B turns positive almost immediately and stays there, while Province A transitions from a neutral warm-up phase to consistently positive lift.}
    \label{fig:monthly_lift}
\end{figure}

Figure~\ref{fig:monthly_lift} presents the monthly relative lift, providing a granular view of the feedback loop's temporal evolution.

\begin{table}[t]
\centering
\caption{Monthly conversion breakdown. Values are new conversions per month. June 2025 is partial (through June 4, the end of the 160-day test window), hence the lower counts.}
\label{tab:monthly}
\small
\begin{tabular}{lcccccc}
\toprule
& \multicolumn{3}{c}{\textbf{Province A}} & \multicolumn{3}{c}{\textbf{Province B}} \\
\cmidrule(lr){2-4} \cmidrule(lr){5-7}
\textbf{Month} & \textbf{Treat} & \textbf{Ctrl} & \textbf{Lift} & \textbf{Treat} & \textbf{Ctrl} & \textbf{Lift} \\
\midrule
Dec '25 & 93 & 101 & $-$7.9\% & 112 & 113 & $-$0.9\% \\
Jan '26 & 285 & 295 & $-$3.4\% & 350 & 311 & +12.5\% \\
Feb '26 & 246 & 247 & $-$0.4\% & 294 & 300 & $-$2.0\% \\
Mar '26 & 394 & 346 & +13.9\% & 459 & 448 & +2.5\% \\
Apr '26 & 485 & 483 & +0.4\% & 611 & 550 & +11.1\% \\
May '26 & 578 & 517 & +11.8\% & 767 & 667 & +15.0\% \\
Jun '26 & 61 & 57 & +7.0\% & 60 & 52 & +15.4\% \\
\bottomrule
\end{tabular}
\end{table}

Table~\ref{tab:monthly} reveals several important patterns:

\begin{itemize}
    \item \textbf{Warm-up duration varies by market}: Province A's warm-up persists through February (three months of neutral-to-slightly-negative lift), while Province B reaches positive lift as early as January (+12.5\%) despite a brief dip in February.
    \item \textbf{Lift magnitude grows over time}: Both provinces show their strongest lifts in the later months (May--June), confirming the compounding nature of the feedback loop. As each iteration improves ranking quality, subsequent iterations benefit from higher-quality deployment data.
    \item \textbf{Non-monotonic but trending positive}: The month-to-month lift is not perfectly monotonic (e.g., Province A shows +13.9\% in March but only +0.4\% in April), reflecting natural market fluctuations and the monthly granularity of model updates.
\end{itemize}

\subsubsection{Cumulative Gap Analysis}

\begin{figure}[t]
    \centering
    \includegraphics[width=\columnwidth]{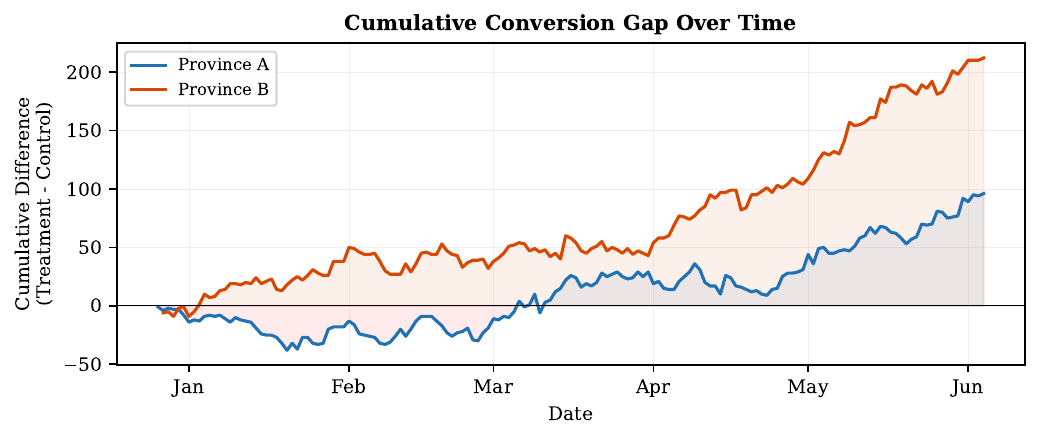}
    \caption{Cumulative conversion difference (Treatment $-$ Control) over time. Both curves show a clear regime change from negative/flat to persistently positive and growing.}
    \label{fig:gap}
\end{figure}

Figure~\ref{fig:gap} visualizes the treatment-control gap over time for both provinces simultaneously, providing the clearest evidence of the feedback loop's sustained effect.

Both provinces converge to a clear \textbf{divergence regime} in which the gap grows approximately linearly (Province A: $\sim$1.0 additional conversions/day; Province B: $\sim$1.4/day), indicating stable per-iteration improvement rather than diminishing returns within the observation window.
Province B enters this regime almost immediately, whereas Province A first passes through a brief warm-up before the gap crosses zero and begins to grow.

This roughly linear growth suggests that, within the observed horizon, the feedback loop has not yet saturated---further iterations would likely continue to improve performance, consistent with the argument for continuous online learning in non-stationary environments.

\subsection{Ablation Study}
\label{sec:ablation}

We analyze the contribution of each component in Table~\ref{tab:ablation}.
All variants are evaluated offline after 4 iterations, so that each component's effect is measured once the feedback loop has had several rounds to take effect.

\begin{table*}[t]
\centering
\caption{Ablation study on reward and loss components (offline evaluation after 4 iterations).}
\label{tab:ablation}
\small
\begin{tabular}{lccccc}
\toprule
\textbf{Variant} & \textbf{P@K} & \textbf{R@K} & \textbf{NDCG@K} & \textbf{AUC} & \textbf{$\Delta$ P@K} \\
\midrule
\method{} (Full) & \textbf{7.56\%} & \textbf{8.40\%} & \textbf{0.4012} & \textbf{0.7891} & --- \\
\midrule
w/o Position $g(r)$ & 7.18\% & 7.98\% & 0.3814 & 0.7853 & $-5.0\%$ \\
w/o Velocity $v(\tau)$ & 7.41\% & 8.23\% & 0.3936 & 0.7871 & $-2.0\%$ \\
w/o Listwise $\LL_{\text{list}}$ & 6.89\% & 7.66\% & 0.3621 & 0.7812 & $-8.9\%$ \\
Pointwise Only (BCE) & 6.61\% & 7.35\% & 0.3548 & 0.7784 & $-12.6\%$ \\
w/o BCE Regularization & 7.29\% & 8.10\% & 0.3889 & 0.7724 & $-3.6\%$ \\
\bottomrule
\end{tabular}
\end{table*}

\paragraph{Listwise loss is the most critical component.}
Removing $\LL_{\text{list}}$ causes the largest single-component drop ($-8.9\%$ P@K), confirming that group-relative optimization is essential.
The ``Pointwise Only'' variant, which uses online deployment data but trains with BCE loss alone, performs only marginally better than the SFT baseline (6.61\% vs.~6.37\%).
This demonstrates that access to online data alone is insufficient; the listwise objective with performance-aware rewards is what drives the improvement.

\paragraph{Position gain provides substantial benefit.}
Removing $g(r)$ reduces P@K by $-5.0\%$.
Position feedback directly encodes ranking quality from deployment: the model learns that correctly ranking a conversion at position 50 is more valuable than at position 5{,}000.
This signal is unique to the online setting and unavailable in offline training.

\paragraph{BCE regularization prevents calibration loss.}
Removing the BCE regularization term reduces P@K by $-3.6\%$ and notably drops AUC by $-2.1\%$ (from 0.789 to 0.772).
This indicates that the listwise loss alone, while optimizing ranking quality, can cause the model to lose calibration on individual predictions.
The regularization anchors pointwise predictions while the listwise term optimizes relative ordering.

\paragraph{Velocity provides incremental gains.}
Removing $v(\tau)$ reduces P@K by $-2.0\%$.
Faster conversions signal stronger purchase intent; this auxiliary signal refines the model's quality estimation beyond binary conversion labels.

\subsection{Ranking Quality Validation with Deployment Metrics}
\label{sec:ranking_quality}

The A/B test in \S\ref{sec:online} quantifies the \textit{relative} gain of the online feedback loop over a static baseline. Here we complement it with an \textit{absolute} assessment of ranking quality and business value in production.
Concretely, we analyze the base ranker $f_{\theta_0}$---the supervised backbone from which \method{} is initialized, prior to any online iteration---using production data from 103 days of operation (September 20 -- December 31, 2025).
This isolates the contribution of the ranking architecture itself: any lift the online loop adds (\S\ref{sec:online}) is stacked \textit{on top of} the strong absolute performance established here.

\paragraph{Top-10\% Recall Analysis.}
We evaluate how effectively the model's top-ranked leads capture actual conversions.
Specifically, we compute the 7-day and 14-day recall of the Top-10\% ranked leads (i.e., the fraction of all conversions that fall within the model's top decile).
Table~\ref{tab:recall} reports summary statistics.

\begin{table}[t]
\centering
\caption{Top-10\% recall rates of the base ranker $f_{\theta_0}$ over 103 days of production operation. Recall = conversions in Top-10\% / total conversions.}
\label{tab:recall}
\small
\begin{tabular}{lccc}
\toprule
\textbf{Window} & \textbf{Mean} & \textbf{Std} & \textbf{Range} \\
\midrule
7-day recall & 44.1\% & 2.4\% & 37.6--48.7\% \\
14-day recall & 41.3\% & 2.1\% & 35.7--45.3\% \\
30-day recall & 35.8\% & 3.7\% & 24.5--40.2\% \\
\bottomrule
\end{tabular}
\end{table}

The model consistently captures $\sim$44\% of 7-day conversions and $\sim$41\% of 14-day conversions within just the top decile of ranked leads.
This demonstrates strong ranking quality even before the online loop is engaged: a random ranker would achieve 10\% recall at the 10\% cutoff, while the base ranker achieves 4.4$\times$ lift over random for 7-day conversions.

\paragraph{Incremental High-Intent Identification.}
A key business application of \method{} is identifying high-intent leads that sales specialists have not yet recognized.
We define \textit{incremental leads} as those ranked high-intent by the model but \textit{not} previously flagged by specialists as likely-to-convert (3/7-day intent markers).

Over the same 103-day production window:
\begin{itemize}
    \item The model identifies $\sim$4,600 incremental high-intent leads per day.
    \item These incremental leads achieve a \textbf{14-day conversion rate of 4.66\%}, compared to 2.05\% for the specialist-flagged baseline---a \textbf{2.3$\times$ improvement}.
    \item Cumulative incremental conversions attributed to the model: \textbf{7,220 lock-ins}.
\end{itemize}

These 7,220 lock-ins are especially consequential in our setting: each conversion corresponds to a high-value durable good, so incremental lock-ins that would otherwise have been overlooked translate directly into substantial revenue.
More broadly, this validates that \method{}'s ranking captures genuine purchase intent signals from dialogue understanding that complement human judgment, rather than merely replicating existing specialist labels.

\paragraph{Temporal Stability.}
Figure~\ref{fig:recall_stability} in Appendix shows that the Top-10\% 7-day recall remains stable between 40--48\% throughout the 103-day observation, with no degradation trend.
This confirms that the ranking backbone delivers consistent quality under real production traffic---a prerequisite for the online loop to build upon reliably in long-term deployment.

\subsection{Sensitivity Analysis}
\label{sec:sensitivity}

We examine how \method{}'s advantage varies with the operational capacity $K$, which determines the size of the daily lead pool distributed to sales specialists.
Table~\ref{tab:k_sensitivity} shows P@K lift over LLM+DPO at different $K$ values.

\begin{table}[t]
\centering
\caption{P@K relative lift of \method{} over LLM+DPO at different $K$.}
\label{tab:k_sensitivity}
\small
\begin{tabular}{lcc}
\toprule
\textbf{$K$} & \textbf{Relative Lift} & \textbf{Remark} \\
\midrule
1K & +21.4\% & strongest gain \\
5K & +18.7\% & high-value window \\
10K & +15.8\% & default setting \\
20K & +11.2\% & moderate gain \\
50K & +5.3\% & diluted advantage \\
100K & +1.9\% & near full coverage \\
\bottomrule
\end{tabular}
\end{table}

\method{}'s advantage is largest at smaller $K$ values (+21.4\% at $K=1{,}000$) and diminishes as $K$ increases.
This is expected: the listwise optimization with position-weighted rewards primarily improves Top-$K$ ranking quality, which matters most when the operational window is tight.
At large $K$ values approaching full coverage, the ranking advantage naturally diminishes as most convertible leads are already included regardless of ordering.
Note that $K$ is set by operational capacity (here $K=10{,}000$ leads per day) rather than tuned; the analysis shows that \method{}'s benefit would be even larger under tighter capacity, and remains positive across the full range.

\section{Conclusion}
\label{sec:conclusion}

We presented \method{}, an online reinforcement learning framework for lead ranking that addresses three fundamental gaps in existing approaches: the offline-online distribution mismatch, the pointwise-listwise objective misalignment, and the historical-current distribution drift.

Our key contributions include: (1) formalizing these three gaps and proposing a closed-loop paradigm for continuous adaptation; (2) designing a performance-aware reward that incorporates conversion outcomes, ranking position, and conversion velocity; and (3) developing Discriminative GRPO, a listwise optimization objective that adapts group-relative advantages to discriminative ranking models.

Comprehensive experiments on large-scale industrial data (16.5M leads, 160 days, two provincial markets with 280 specialists) demonstrate that \method{} achieves consistent improvements over static baselines.
A production A/B test validates +4.7\% and +8.7\% cumulative lift in lock-in conversions ($p=0.047$ and $p=0.002$ respectively), with the advantage widening as the feedback loop accumulates deployment data---evidence of its compounding nature.
Deployment metrics further validate the absolute ranking quality of the backbone: Top-10\% recall of 44.1\% (4.4$\times$ random) and 7,220 incremental high-value conversions from model-identified high-intent leads that specialists had overlooked.

\paragraph{Limitations and Future Work.}
The current system operates at monthly granularity due to the 30-day conversion window.
Future directions include: (1) \textbf{faster feedback}: leveraging intermediate signals (e.g., test drive bookings, showroom visits) to enable weekly or daily iterations; (2) \textbf{multi-objective optimization}: balancing short-term conversion with long-term customer lifetime value; (3) \textbf{cross-domain transfer}: extending \method{} to other long-cycle sales domains such as real estate and enterprise B2B; and (4) \textbf{exploration-exploitation}: incorporating uncertainty-aware selection to balance ranking quality with data collection for future learning.
Additionally, investigating alternative reward formulations (e.g., DPO-style pairwise preferences derived from performance outcomes) and more sophisticated listwise losses (e.g., NeuralNDCG~\cite{pobrotyn2021neuralndcg}) are promising directions.

\bibliographystyle{plain}
\bibliography{references}

\newpage
\appendix
\section{Additional Online Deployment Analysis}
\label{sec:appendix}

This appendix provides supplementary analyses of the online deployment data to further characterize \method{}'s feedback loop dynamics.

\subsection{Daily Conversion Rate Analysis}

\begin{figure*}[t]
    \centering
    \includegraphics[width=\textwidth]{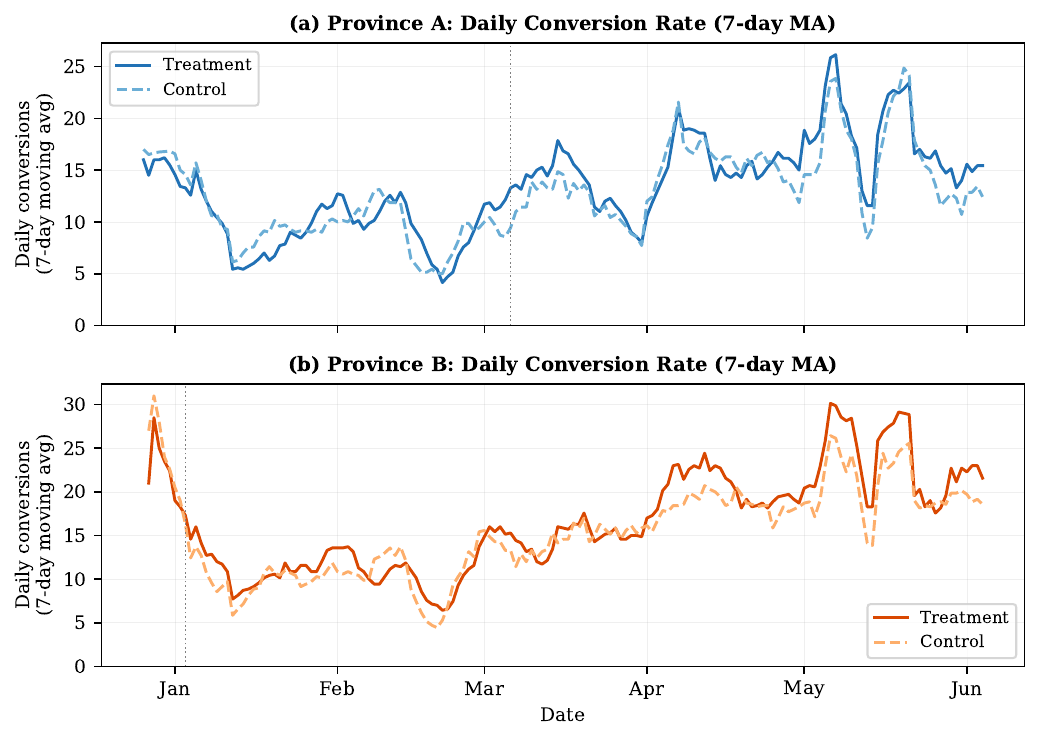}
    \caption{Daily new conversions (7-day moving average) for treatment and control groups. Province A shows treatment overtaking control around March 2026; Province B shows treatment consistently above control from early January.}
    \label{fig:daily_rate}
\end{figure*}

Figure~\ref{fig:daily_rate} presents the 7-day moving average of daily new conversions for treatment and control groups in both provinces.
Unlike the cumulative view in Figure~\ref{fig:cumulative}, this visualization reveals the \textit{instantaneous} productivity differences.

Key observations:
\begin{itemize}
    \item \textbf{Province A}: The treatment and control daily rates are nearly identical during December--February, with occasional crossovers. Starting March 2026, the treatment curve consistently exceeds the control, with the gap widening particularly during high-volume periods (March, May).
    \item \textbf{Province B}: The treatment daily rate exceeds the control almost immediately and maintains this advantage throughout, with the gap becoming more pronounced in later months as the feedback loop compounds.
    \item \textbf{Shared seasonal patterns}: Both treatment and control exhibit similar seasonal fluctuations (e.g., dips during Chinese New Year in mid-February, peaks during promotional periods in March and May), confirming that the randomization successfully balanced external factors between groups.
\end{itemize}

\subsection{Per-Specialist Productivity}

\begin{figure}[t]
    \centering
    \includegraphics[width=\columnwidth]{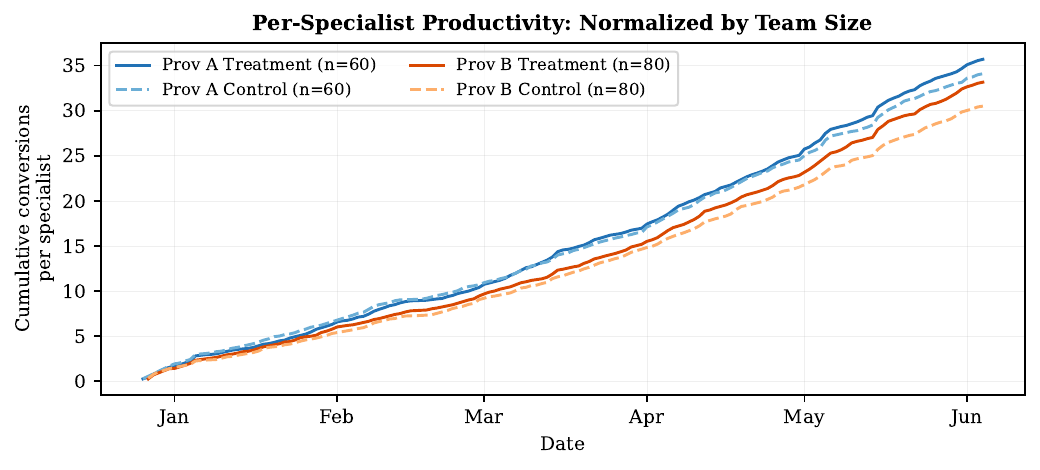}
    \caption{Cumulative conversions per specialist, enabling cross-province comparison on a normalized scale.}
    \label{fig:per_specialist}
\end{figure}

To control for the difference in team sizes between provinces (60 per group in Province A vs.\ 80 in Province B), Figure~\ref{fig:per_specialist} normalizes cumulative conversions by the number of specialists in each group.

After normalization, the per-specialist productivity reveals:
\begin{itemize}
    \item Both provinces have similar per-specialist baselines ($\sim$30--36 conversions per specialist over 160 days).
    \item \method{} provides +1.6 conversions per specialist in Province A and +2.7 in Province B.
    \item The per-specialist curves are closely matched across provinces for the same condition (e.g., both control groups converge to $\sim$30--34/specialist), suggesting comparable market conditions despite different team sizes.
\end{itemize}

These normalized results confirm that the lift observed in aggregate is not an artifact of team size differences and reflects genuine per-specialist productivity gains from improved lead ranking.

\subsection{Ranking Quality Temporal Stability}

\begin{figure}[t]
    \centering
    \includegraphics[width=\columnwidth]{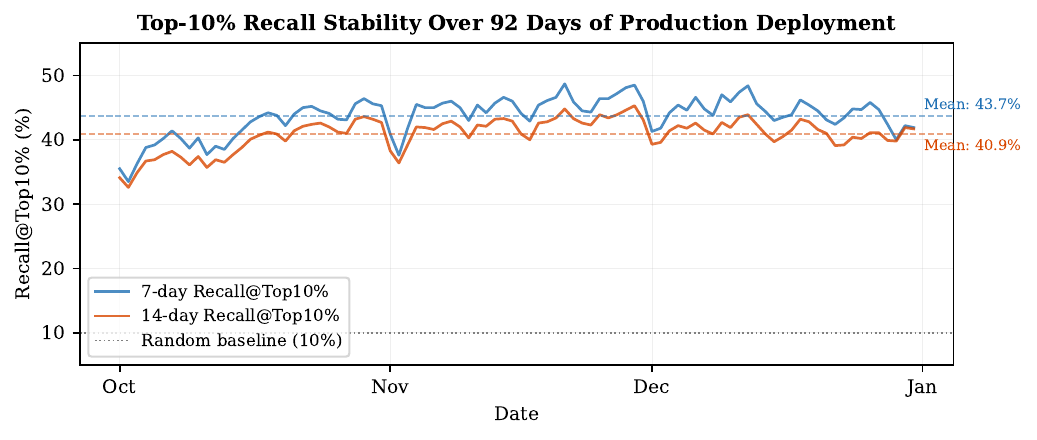}
    \caption{Top-10\% recall stability over 103 days. The base ranker $f_{\theta_0}$ maintains 7-day recall between 37--49\% (mean 44.1\%) and 14-day recall between 36--45\% (mean 41.3\%), with no degradation trend. The random baseline (10\%) is shown for reference.}
    \label{fig:recall_stability}
\end{figure}

Figure~\ref{fig:recall_stability} shows the daily Top-10\% recall rates over the 103-day production window (September 20 -- December 31, 2025).

Key observations:
\begin{itemize}
    \item The 7-day recall remains in the 40--48\% range during stable periods, representing a \textbf{4.4$\times$ lift over random} (10\% baseline).
    \item No downward trend is observed over the 103-day window, indicating that the ranking backbone is robust to day-to-day distribution shifts in production.
    \item Brief dips (e.g., early October, early November) coincide with major product launches that temporarily shift the lead distribution, after which recall recovers.
\end{itemize}

\subsection{Incremental Value Over Human Judgment}

\begin{figure}[t]
    \centering
    \includegraphics[width=0.7\columnwidth]{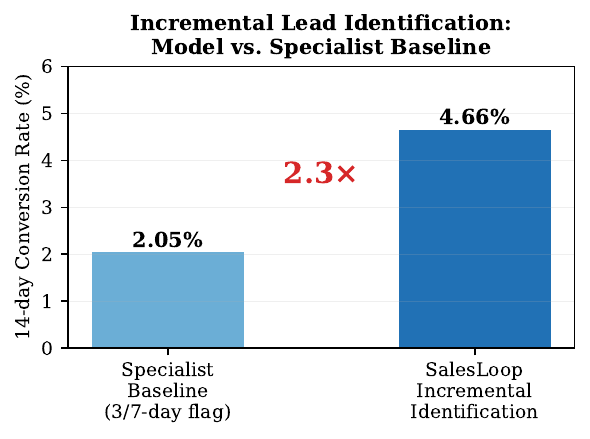}
    \caption{14-day conversion rate comparison: specialist-flagged leads (2.05\%) vs. \method{} incremental identification (4.66\%). The model identifies high-intent leads that specialists miss, at 2.3$\times$ higher conversion rate.}
    \label{fig:incremental}
\end{figure}

Figure~\ref{fig:incremental} compares the conversion rate of \method{}-identified incremental leads versus the specialist-flagged baseline.

This result demonstrates that \method{}'s dialogue-based ranking captures latent purchase intent signals that even experienced sales specialists cannot identify from their standard workflow. The model effectively serves as a \textit{complementary intelligence layer} rather than a replacement for human judgment.

\end{document}